\documentclass{article}

    \usepackage[preprint]{neurips_2023}

\usepackage{hyperref}       
\usepackage{url}            
\usepackage{booktabs}       
\usepackage{amsfonts}       
\usepackage{nicefrac}       
\usepackage{microtype}      
\usepackage{xcolor}         

\usepackage{graphicx}
\usepackage{subfigure}
\usepackage{times}
\usepackage{algorithm}
\usepackage{algorithmic}

\usepackage{latexsym}
\usepackage{amsmath}
\usepackage{amssymb}
\usepackage{enumerate}
\usepackage{enumitem}
\usepackage{inconsolata}
\usepackage{wrapfig,lipsum}
\makeatletter
\renewcommand{\maketag@@@}[1]{\hbox{\m@th\normalfont#1}}%
\makeatother

\newcommand{\name}{SeqDiffuSeq }

\title{SeqDiffuSeq: Text Diffusion with Encoder-Decoder Transformers}

\author{Hongyi Yuan$^{12}$\thanks{$\quad$Work done at Alibaba DAMO Academy.}  , Zheng Yuan$^2$, Chuanqi Tan$^2$, Fei Huang$^2$, Songfang Huang$^2$   \\                         
  $^1$Tsinghua University, $^2$Alibaba Group \\
  \texttt{yuanhy20@mails.tsinghua.edu.cn} \\
  \texttt{\{yuanzheng.yuanzhen,chuanqi.tcq,f.huang,songfang.hsf\}@alibaba-inc.com}
  }
  
\begin{document}

\maketitle

\begin{abstract}
The diffusion model, a new generative modeling paradigm, has achieved great success in image, audio, and video generation.
However, considering the discrete categorical nature of the text, it is not trivial to extend continuous diffusion models to natural language.  
In this work, we propose SeqDiffuSeq, a text diffusion model, to approach sequence-to-sequence text generation with an encoder-decoder Transformer architecture.
To improve the generation performance, SeqDiffuSeq is equipped with the self-conditioning technique and our newly proposed adaptive noise schedule technique. 
Self-conditioning enables SeqDiffuSeq to better use the predicted sequence information during the generation process.
The adaptive noise schedule balances the difficulty of denoising across time steps at the token level.
Experiment results illustrate the improved performance on five sequence-to-sequence generation tasks compared to other diffusion-based models regarding text quality and inference time. 
We have released our codes.
\footnote{\url{https://github.com/Yuanhy1997/SeqDiffuSeq}}
\end{abstract}

\section{Introduction}

Generative modeling is drawing more attention in recent years of machine learning research due to the development of diffusion models \citep{ho2020denoising}. Diffusion models define the forward process and the reverse process where the former gradually diffuses data to random noise while the latter recovers data from random noise iteratively, which have shown superior performance on synthesizing images \citep{Rombach2021HighResolutionIS}, audios \citep{kong2020diffwave}, and videos \citep{Ho2022VideoDM} over other generative methods, such as generative adversarial network (GAN) \citep{gan} and normalizing flow \citep{Kobyzev_2021}.  

It is not trivial to extend diffusion models to the generation of natural languages. Most of the existing diffusion models are applied to continuous feature space \citep{ho2020denoising, nichol2021improved} while texts are sequences of discrete categorical tokens. Recently, research has explored categorical diffusion models in discrete space for text generation \citep{argmaxflowanddiffu,austin-etal-2022-community}. There also exists research such as DiffusionLM \citep{li2022diffusion} that applies continuous diffusion models to word embedding. 
However, these works only focus on unconditional and controlled text generation. 

Sequence-to-sequence text generation is a fundamental natural language processing setting and covers various practical downstream tasks, such as dialogue \citep{Ni2021RecentAI} and machine translation \citep{Liu2020MultilingualDP}.
In recent practice, researchers resort to auto-regressive (AR) \citep{dai-etal-2019-transformer} or non-auto-regressive (NAR) \citep{Gu2019LevenshteinT} Transformers for the tasks, and achieve good generation performance. 
Using diffusion models, a recent work named DiffuSeq \citep{gong2022diffuseq} applies the diffusion-based method for sequence-to-sequence text generation. They deploy encoder-only Transformers and partially define diffusion and denoising processes on output sequences.

In this work, we explore diffusion models with encoder-decoder Transformer architecture for sequence-to-sequence generation. 
We propose \name  which extends the continuous diffusion framework proposed in DiffusionLM \citep{li2022diffusion} to sequence-to-sequence settings. We equip \name with the self-conditioning technique \citep{analogbits} and our newly proposed adaptive noise schedule. 
Self-conditioning helps the model better capture the information from former iterations during the generation.
The proposed adaptive noise schedule learns a token-level noise schedule to better control the amount of noise injected and information recovered during the forward and reverse process \citep{nichol2021improved}.

We conduct experiments on five generation tasks. Results show that \name achieves competitive performance compared with AR and NAR baselines in terms of generation quality and diversity. \name also shows improved generation performance and inference speed compared to text diffison model DiffuSeq. Ablation studies demonstrate that our model can benefit from self-conditioning and adaptive noise schedule techniques, and both are complementary to each other in sequence-to-sequence settings.

To summarize, the main contributions of this work are as follows:
\begin{enumerate}
\item We propose \name that extends the continuous text diffusion model to sequence-to-sequence text generation with encoder-decoder Transformer architecture.

\item The self-conditioning and newly proposed adaptive noise schedule technique can effectively improve the generation quality of the text diffusion model.
    
\item Experiments show \name achieves promising performance with the previous diffusion-based method DiffuSeq as well as AR and NAR models on five generation tasks. 
\end{enumerate}

\section{Related Work}


Since the great success of diffusion models in vision \citep{ho2020denoising,Rombach2021HighResolutionIS,song2021scorebased}, researchers have explored extending diffusion models to text generation. Considering the discrete and categorical nature of texts, Multinomial Diffusion \citep{argmaxflowanddiffu} and D3PM \citep{austin2021structured} are proposed for modeling categorical data. They define discrete diffusion models using discrete categorical transitions directly on texts.  DiffusionBERT \citep{he2022diffusionbert} follows D3PM and introduces pre-trained models for language modeling. Besides, recent research also explores converting texts into continuous features to adapt to diffusion models. Bit Diffusion \citep{analogbits} encodes discrete data as binary bits and treats these binary bits as real number features. \citet{diffusionebm} is proposed to build text diffusion models in continuous latent space. DiffusionLM \citep{li2022diffusion} uses the word embedding space for continuous diffusion models and introduces auxiliary losses to enable joint learning of embedding and network parameters. Following DiffusionLM, recent research explores improving text generation quality \citep{Strudel2022SelfconditionedED}, and DiffuSeq \citep{gong2022diffuseq} extends it to sequence-to-sequence settings. Compared to DiffuSeq, we propose a different model architecture and self-conditioning and adaptive noise schedule techniques to improve sequence-to-sequence generation performance.

Noise schedules in diffusion models control the level of noise injected and the level of information recovered in the forward and reverse process respectively. Previous research in vision and texts demonstrates that appropriate noise schedule design can improve the generation quality performance of diffusion models \citep{nichol2021improved,li2022diffusion}. Concurrently, DiffusionBERT \citep{he2022diffusionbert} proposes a spindle schedule for language modeling, and CDCD \citep{dieleman2022continuous} designs a learned noise schedule for language modeling and machine translation. Different from both concurrent works, SeqDiffuSeq is proposed with a token-level noise schedule that balances the difficulty of denoising across time steps. \citet{gao2023difformer} proposes Difformer and is orthogonal to our work.

\section{Preliminary}
\label{preliminary}

Diffusion model is generally formulated by a designed forward diffusion process and a learnt reverse denoising process. In the forward diffusion process, samples gradually mix with random noise, while in the reverse denoising process, the random noise is gradually denoised to generate synthetic samples. 
In this work, our diffusion model adopts the forward and reverse processes proposed in DDPM \citep{ho2020denoising}. 

For the forward process, given a sample $z_0$ from a real-world data distribution $q(z_0)$. At each time step $t\in\{1,2,\cdots,T\}$, a noise sample $z_t$ is sampled from $z_t\sim q(z_t|z_{t-1})=\mathcal{N}(z_t;\sqrt{\alpha_t}z_{t-1}, (1-\alpha_t)I)$, where $\alpha_t$ control the noise added at time step $t$. 
In this regard, when $T$ is large enough, a real-world sample will gradually and ultimately diffuse to a standard Gaussian noise distribution. 

For the reverse process, the diffusion model uses a learnt parameterized denoising distribution $z_{t-1}\sim p_\theta(z_{t-1}|z_t)$ to gradually recover samples from noise. 
The denoising distribution is parameterized by $\theta$ and is to fit the posterior distribution $q(z_{t-1}|z_t,z_0)$ of the forward process.
$q(z_{t-1}|z_t,z_0)$ can be derived as:
\begin{align}
    &q(z_{t-1}|z_t,z_0)=\mathcal{N}(z_{t-1};\tilde{\mu}(z_0,z_{t}), \tilde{\beta}_{t}I). \label{equ:posterior}
\end{align}
In this equation, 

\begin{align}
    \tilde{\mu}(z_0,z_{t})=\frac{\sqrt{\bar{\alpha}_{t-1}}\beta_{t}}{1-\bar{\alpha}_{t}}{z_0}+\frac{\sqrt{\alpha_{t}}(1-\bar{\alpha}_{t-1})}{1-\bar{\alpha}_{t}}z_{t}, \label{equ:posterior_mean}\\
\bar{\alpha}_t=\prod_{s=1}^t\alpha_s, \quad \beta_t=1-\alpha_t, \quad \tilde{\beta}_{t} = \frac{1-\bar{\alpha}_{t-1}}{1-\bar{\alpha}_t}\beta_t.
\end{align}

With learnt denoising distribution $p_\theta$, a synthetic real-world sample $z_0$ can be generated from pure random noise $z_T$ step-by-step.

\begin{figure*}
    \centering
    \resizebox{1.\columnwidth}{!}{
    \includegraphics{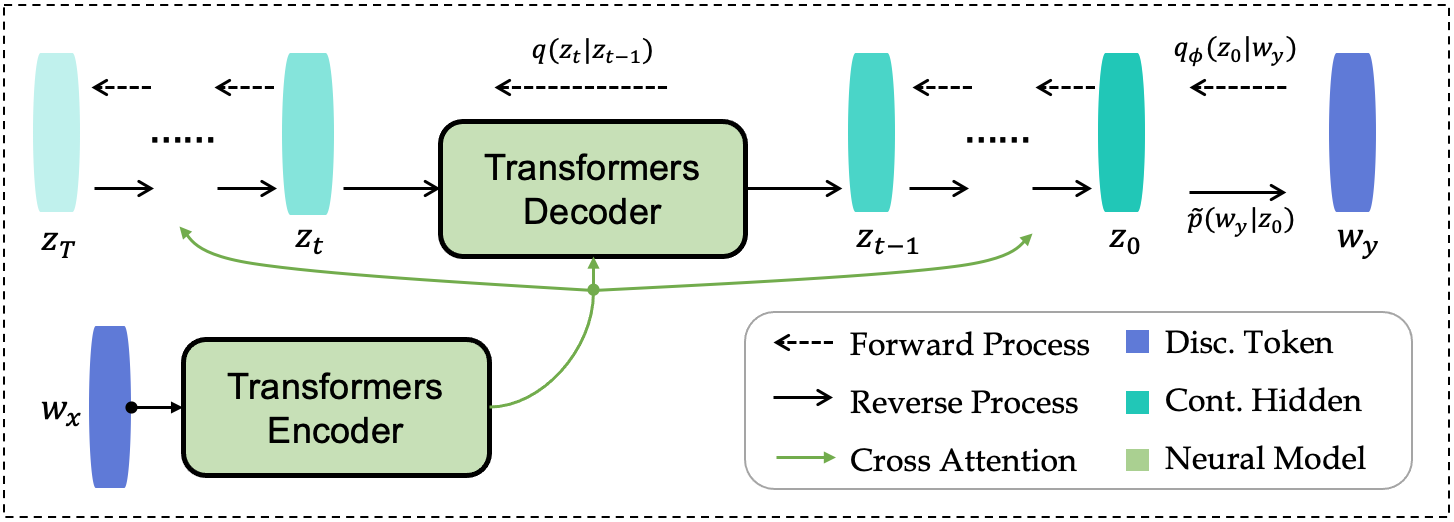}
    }
    \caption{The overview of SeqDiffuSeq with an encoder-decoder Transformers architecture.
    }
    \label{fig:overview}
\end{figure*}

\section{Approach}
\label{approach}

In this section, we present the main design of our proposed SeqDiffuSeq for sequence-to-sequence language generation. The overview of SeqDiffuSeq is depicted in Figure \ref{fig:overview}.
In the following sections, the input and output sequences are denoted as $w_x$ and $w_y$ respectively. For the $i$-th token in $w_y$, the token is denoted as $w_y^i$, where $0<i\le n$ and $n$ represents the maximum output sequence word length. In order to avoid lengthy notations, we omit the indices referring to different data samples.

\subsection{Diffusion Model}

\paragraph{Forward Process} 
To fit diffusion models to sequence-to-sequence settings, we extend the text diffusion model, DiffusionLM \citep{li2022diffusion}. 

In the sequence-to-sequence setting, the forward process gradually changes the target output sequence $w_y$ to random noise. Diffusing $w_y$ to pure random noise is independent of the input sequence $w_x$.
For the sequence $w_y$, we use an embedding function $g_\phi$ to map the word tokens $w^i_y$ to continuous word embedding $g_\phi(w^i_y)\in \mathbb{R}^d$, where $d$ represents the dimension of embedding and $\phi$ represents the parameters of the word embedding function. The embedding for the sequence $w_y$ is defined by stacking the tokens' embedding and is denoted as $g_\phi(w_y)\in \mathbb{R}^{n\times d}$. At the beginning of the forward process, a Markovian transition parameterized by $q_\phi (z_0|w_y) = \mathcal{N}(z_0;g_\phi(w_y), \beta_0 I)$ is added. 
Extended by $q_\phi (z_0|w_y)$, the forward process can continue to diffuse continuous features of $z_0$ iteratively.
For each time step $t$, we apply the diffusion distribution $q(z_t|z_{t-1})$ to get noisier samples. Ultimately, the output sequence $w_y$ becomes $z_T$ which is nearly pure random noise following standard Gaussian distribution. 

\paragraph{Reverse Process} Diffusion models generate the synthetic samples by successively sampling the denoising distribution in the reverse process. For each time step $t$ in the reverse process, a learnt denoising distribution $p_\theta$ parameterized by $\theta$ generates samples $z_{t-1}$ conditioned on the former noisier samples $z_t$. In the sequence-to-sequence setting, the generated sequences correlate to input sequences. Therefore, the denoising distribution is additionally conditioned on the input sequence $w_x$, and $p_\theta=p_\theta(z_{t-1}|z_t, w_x)$.
After the reverse denoising process reaches $T=0$, we round each column of the generated $\hat{z}_0$ to its nearest word in the embedding space by the rounding distribution $\tilde{p}_\phi(w_y|\hat{z}_0)$ to generate the final word sequences.

\paragraph{Training Objective} 
We optimize $\theta$ and embedding parameters by minimizing the variational bound of the data log-likelihood:
\begin{align}
   & \mathcal{L}_{VB} = \mathbb{E}_{q_\phi(z_{0:T}, w_x,w_y)}\bigg[\log{\frac{q(z_T|z_0)}{p(z_T)}}  +\sum_{t=2}^T\log{\frac{q(z_{t-1}|z_0,z_t)}{p_\theta(z_{t-1}|z_t,w_x)}}\nonumber\\
   &-\log{p_\theta(z_0|z_1,w_x)} +\log{q_\phi(z_0|w_y)}-\log{\tilde{p}_\phi(w_y|z_0)}\bigg] ,\label{app:loss_vb}
\end{align}
The training objective is to narrow down the discrepancy between $p_\theta(z_{t-1}|z_t, w_x)$ and the posterior $q(z_{t-1}|z_t,z_0)$ in the forward process. Since $q(z_{t-1}|z_t,z_0)$ follows the form of Gaussian distribution, we parameterize the denoising distribution following Gaussian distribution family and 
$p_\theta(z_{t-1}|z_{t}, w_x) = \mathcal{N}(z_{t-1}; \tilde{\mu}_\theta(z_{t}, w_x, t), \tilde{\beta}_{t}I),$
where
\begin{align}
    &\tilde{\mu}_\theta(z_{t}, w_x, t) = \frac{\sqrt{\bar{\alpha}_{t-1}}\beta_{t}}{1-\bar{\alpha}_{t}}z^0_\theta(z_{t}, w_x, t)+\frac{\sqrt{\alpha_{t}}(1-\bar{\alpha}_{t-1})}{1-\bar{\alpha}_{t}}z_{t}.\label{equ:denoise_mean}
\end{align}
$z^0_\theta(z_{t}, w_x, t)$ is named the denoising function and predicts the estimated output embedding sequences at each reverse step $t$.
%
%
Then according to density functions $q$ and $p_\theta$ following Gaussian distribution, the objective can be further simplified as:
\begin{align}
    &\mathcal{L}_{simple} = \mathbb{E}_{q_\phi(z_0,w_x,w_y)}\bigg[\sum_{t=2}^T\mathbb{E}_{q(z_t|z_0)} \|z^0_\theta(z_{t}, w_x, t)-z_0\|^2\nonumber\\
    &+ \|\mu_T(z_0)\|^2+\|z^0_\theta(z_{1}, w_x, 1)-g_\phi(w_y)\|^2-\log{\tilde{p}_\phi(w_y|z_0)}\bigg],\label{equ:simple} 
\end{align}
where $q(z_t|z_0)=\mathcal{N}(z_t;\sqrt{\bar{\alpha}_t}z_0, (1-\bar{\alpha}_t)I)$ for efficient sampling of $z_t$ during training, and $\mu_T(z_0) = \sqrt{\bar{\alpha}_T}z_0$. We leave the detailed derivation to Appendix \ref{app:derive_obj}. The training objective becomes to fit $g_\phi(w_y)$ and the denoising function $z^0_\theta(z_{t}, w_x, t)$, which we can model with encoder-decoder Transformers architectures.
During training, the sampling distribution $q_\phi$ contains trainable parameters of word embedding. We can backpropagate through this with reparameterization trick \citep{kingma2013auto}. 

\paragraph{Denoising with Encoder-Decoder Framework} Unlike DiffuSeq \citep{gong2022diffuseq} using encoder-only Transformer architecture, we propose using an encoder-decoder Transformers architecture to model the input and output text sequences. For $z^0_\theta(z_{t}, w_x, t)$, we use the encoder to process the input sequences $w_x$ and use the decoder to model the noisy output sequence $z_{t}$. Following the previous work \citep{li2022diffusion}, we inject time step information $t$ by adding time step embedding to $z_{t}$. 
Using the encoder-decoder architecture has computational convenience during generation because the input sequences $w_x$ only require one forward computation through the encoder network during the whole reverse process. 
Considering the reverse process requires thousands of iterations to generate the output sequences of high quality, the saving of computational resources can be significant. 

During training and generation, the function $z^0_\theta$ generates denoised samples at the sequence level. Therefore making predictions from the denoising function $z^0_\theta$ resembles the non-autoregressive natural language generation. In this regard, we use a decoder with full attention matrices instead of causal attention matrices to model $z_{t}$ at the sequence level.

\subsection{Self-Conditioning}

At each time step $t$ in the reverse process, the denoising function $z^0_\theta(z_{t}, w_x, t)$ makes output sequence predictions based on the noisier sample $z_t$. $z_t$ is sampled from the former denoising distribution by mixing former sequence prediction $\hat{z}_0^{t} = z^0_\theta(z_{t+1}, w_x, t+1)$, $z_{t+1}$ and random noise. In this regard, part of the information contained in the former prediction $\hat{z}_0^{t}$ is discarded. Bit-Diffusion \citep{analogbits} proposed the self-conditioning technique mitigating this waste of information by additionally taking former sequence predictions as inputs. The denoising function is formulated as $z^0_\theta(z_{t},\hat{z}_0^{t}, w_x, t)$. Self-conditioning may enable the denoising function to refine the former sequence predictions rather than make new predictions from scratch. 
It is empirically verified that the self-conditioning technique can boost the performance of text diffusion models \citep{Strudel2022SelfconditionedED}.

To fit the technique into the Transformers modeling of $z^0_\theta$ in our sequence-to-sequence setting, the sequence features $\hat{z}_0^{t}$ from the former predictions are concatenated with noisier sequence features $z_{t}$ on the embedding dimension. Hence, the dimension of input features of Transformer decoder becomes $n\times 2d$. Since the former sequences at time step $t$ are sampled successively from $T$ to $t$ which is computational-tedious during training, we take an efficient training scheme. With half probability, $z^0_\theta(z_{t},\hat{z}_0^{t}, w_x, t)$ is trained by setting the input $\hat{z}_0^{t}$ to $0$. Otherwise, $\hat{z}_0^{t}$ is first estimated by $z^0_\theta(z_{t},0, w_x, t)$ and then is used for self-conditioning training. Under the second circumstance, we do not backpropagate through the first forward propagate estimated $\hat{z}_0^{t}$. 
\subsection{Adaptive Noise Schedule}


In the domain of vision and audio, the generated sample quality \citep{nichol2021improved} and likelihood estimation \citep{kingma2021on} may potentially benefit from different appropriate time schedules. 
Previous research uses different simple functions such as linear function \citep{ho2020denoising} or cosine function \citep{nichol2021improved} of $\alpha$ against time step $t$ to design noise schedules. Such designs may results in unbalanced denoising difficulties for each step and lead to unsatisfying generation quality. Some works proposed to alleviate this problem by importance sampling \citep{li2022diffusion} or loss reweighing \citep{gong2022diffuseq}. 

We propose a novel adaptive noise schedule at the token-level. Firstly, 
we propose to adaptively adjust the time schedules during training to make the denoising difficulties of $z^0_\theta$ predicting output sequence increase linearly with respect to the time step. 
Secondly, we separately set adaptive noise schedule for different token positions, unlike previous text diffusion research that only designs noise schedules on the whole sequence level. Since the intrinsic features for embedding sequences are different across token positions within, we assume that for different token positions the expected noise schedules are different. 

Concretely, we measure the difficulties of denoising task at each time step $t$ and token position $i$ by the training losses $\mathcal{L}_t^i=\mathbb{E}_{q_\phi (w_x,w_y,z_t,z_0)}\|z^0_\theta(z_t, \hat{z}_0^{t}, w_x, t)^i-z_0^i\|^2$. We use the schedule of $\bar{\alpha}_t^i$ which ranges from 0 to 1 to access the noise schedule design. $\bar{\alpha}_t^i$ controls the noise level at each time step $t$. Our adaptive noise schedule for each token position $i$ is to fit a mapping $\bar{\alpha}^i = M_i(\mathcal{L}^i)$ between $\mathcal{L}_t^i$ and $\bar{\alpha}_t^i$ by linear interpolation. For time step $t$, $\forall x\in [\mathcal{L}_{t-1}^i, \mathcal{L}_t^i)$,
\begin{align}
    M_i(x) = \frac{\bar{\alpha}_t^i-\bar{\alpha}_{t-1}^i}{\mathcal{L}_t^i-\mathcal{L}_{t-1}^i}(x-\mathcal{L}_{t-1}^i) + \bar{\alpha}_{t-1}^i, \label{equ:interp} 
\end{align}
After initializing a noise schedule, 
we record the loss $\mathcal{L}_t^i$\footnote{We do not record the losses $\mathcal{L}_{t}^i$ for the padding tokens.}.
The mapping $M_i$ is fitted 
after each training period. 
Ideally, the training losses should be monotonic with respect to the time step $t$ since for larger $T$ the input features $z_t$ to the denoising function are noisier. 
However, overall time step $T$ is usually by thousands, hence this results in a fine-grained discretization of $\bar{\alpha}^i$.  Due to the empirical loss estimation errors, training losses may not be monotonic between some successive time steps. To alleviate this issue and fit a smoother mapping $M_i$, we form a coarse-grained discretization $s$ for  $\bar{\alpha}^i$ and $\mathcal{L}^i$:
$\mathcal{L}_s^i = \frac{1}{K}\sum_{t=s\times K}^{s\times(K+1)}\mathcal{L}_t^i$, $\bar{\alpha}_s^i = \frac{1}{K}\sum_{t=s\times K}^{s\times(M+1)}\bar{\alpha}_t^i$, $s=\left\lfloor \frac{t}{K} \right\rfloor$,
where $K$ is the stride to evenly downsample $t$ and $\lfloor \cdot \rfloor$ rounds the number down to it nearest integer.

\begin{algorithm}[t] 
\caption{Adaptive Noise Schedule} 
\label{alg:Framwork} 
\begin{algorithmic}[1] 
\ENSURE Current recorded losses $\mathcal{L}_t^i$ and noise schedules $\bar{\alpha}_t^i$ for each time step $t$ and token position $i$
\IF{Train Step \% Update Step == 0}
\FOR{each token position $i$}
\STATE Fit the mapping $M_i$ by Equation \ref{equ:interp},
\STATE Take new $\mathcal{L}_t^{i,new}$ value with equal interval 
between $\min_t(\mathcal{L}_t^i)$ and $\max_t(\mathcal{L}_t^i)$,
\STATE Get new schedule $\bar{\alpha}_t^{i,new}=M_i(\mathcal{L}_t^{i,new})$,
\ENDFOR
\ENDIF
\RETURN Noise schedule $\bar{\alpha}_t^{i,new}$ for each $t$ and $i$
\end{algorithmic}
\end{algorithm}

With the learnt linear interpolation mapping $\bar{\alpha}_s^i = M_i(\mathcal{L}_s^i)$, 
we can obtain the adjusted discretized noise schedule $\bar{\alpha}_{t}^{i,new}$ by $\bar{\alpha}_{t}^{i,new}=M_i(\mathcal{L}_{t}^{i,new})$ where $\mathcal{L}_{t}^{i,new}$'s are evenly taken between the minimum and maximum recorded values.
As the training progresses, we adaptively calibrate the noise schedule $\bar{\alpha}^i$ by repeating the above-mentioned procedure once per training updates. The pseudo-code for setting adaptive noise schedules during training is shown in Algorithm \ref{alg:Framwork}. 

\section{Experiments}

\subsection{Datasets}

We conduct experiments on six datasets across five different text generation tasks: Quora Question Pairs (QQP) \citep{quora-question-pairs} for Paraphrase Generation, Wiki-Auto \citep{jiang2020neural} for Text Simplification, Quasar-T \citep{dhingra2017quasar} for Question Generation, Commonsense Conversation Dataset (CCD) \citep{zhou2018commonsense} for Dialogue Generation as well as the German(DE)-English(EN) pairs of IWSLT14 and WMT14 for Machine Translation. Detailed introductions and statistics of the datasets as shown in Appendix \ref{app:data_stat}. 

\begin{table*}[t]
    \caption{Main results on Paraphrase, Text Simplification, Question Generation, Dialogue, and Machine Translation. We use the results reported in the DiffuSeq paper for CCD results since reproducing CCD results requires more than 10 days of training on 8 NVIDIA A100 80GB GPUs.}
    \label{tab:main-result}
     \centering
     \resizebox{1.0\columnwidth}{!}{
    \begin{tabular}{l|cccccc}
    \hline
    \hline
     &\multicolumn{3}{c}{\textbf{QQP}}&\multicolumn{3}{c}{\textbf{Wiki-Auto}}  \\
     &BLEU &BERTScore &dist. 1 & BLEU &BERTScore & dist. 1\\
    \hline
    Transformers&5.80&53.92&78.89&24.45&75.90&88.86\\
    GPT2-large FT&20.59&83.63&98.19&26.93&78.82&94.64\\
    LevT &22.68&83.44&97.90&20.52&72.54&97.15\\
    DiffuSeq &18.47&79.47&97.61 &29.89&79.12&92.33 \\
    DiffuSeq w/ MBR=10&24.13&83.65&98.07&36.43&81.39&92.61 \\
    \hline
    \textbf{SeqDiffuSeq}&23.28&82.91&98.06&37.09&82.11&90.81 \\
    \textbf{SeqDiffuSeq} w/ MBR=10 &24.34&84.00&98.07&37.12&82.14&90.77\\
     \hline\hline
     &\multicolumn{3}{c}{\textbf{Quasar-T}} &\multicolumn{3}{c}{\textbf{CCD}}\\
     &BLEU &BERTScore &dist. 1 & BLEU &BERTScore & dist. 1\\
      \hline
    Transformers&3.64&53.34&82.36&1.89&47.81&74.93\\
    GPT2-large FT&11.10&63.46&96.70&1.25&52.93&92.44\\
    LevT &9.30&54.91&89.14&1.58&47.60&97.26 \\
    DiffuSeq &15.84&59.39&91.12&-&-&-  \\
    DiffuSeq w/ MBR=10&17.01&60.95&90.72&1.39&51.31&94.67  \\
    \hline
   \textbf{SeqDiffuSeq} &17.20&61.35&92.70&0.84&43.82&96.50\\
    \textbf{SeqDiffuSeq} w/ MBR=10&17.46&61.74&92.48&1.12&44.25&96.08 \\
    \hline
    \hline
    &\multicolumn{2}{c}{\textbf{IWSLT14}} &\multicolumn{4}{c}{\textbf{WMT14}}  \\
     &\multicolumn{1}{c}{EN-DE}&\multicolumn{1}{c}{DE-EN}&\multicolumn{2}{c}{EN-DE}&\multicolumn{2}{c}{DE-EN}\\
     &SacreBLEU &SacreBLEU &SacreBLEU & BLEU &SacreBLEU & BLEU\\
     \hline
     Transformers &26.51&33.81&26.20&27.48 &30.20&31.19\\
     CMLM w/ iter=1 &14.36&21.46&-&18.05&-&21.83\\
     CMLM w/ iter=4 &23.74&32.83&-&25.94&-&29.90\\
     CDCD&-&-&19.30&-&24.90&-\\
     CDCD w/ MBR=10&-&-&19.70&-&25.40&-\\
     \hline
     \textbf{SeqDiffuSeq} & 21.96&30.16&19.16&23.63&23.28&25.22\\
     \textbf{SeqDiffuSeq} w/ MBR=10 &22.12&30.45&19.76&24.24&23.93&25.90\\
     \hline \hline
    \end{tabular}
    }
\end{table*}

\subsection{Baselines}

We consider three kinds of models as baselines. 
First, vanilla encoder-decoder Transformers and pre-trained GPT-2 are selected as strong AR baselines.
Second, since \name denoises outputs at the sequence level, we compare it with an NAR baseline Levenshtein Transformer (LevT) \citep{Gu2019LevenshteinT}. For machine translation, we also use CMLM \citep{ghazvininejad-etal-2019-mask} which is an NAR translation method with iterative refinement as baselines.
Besides, we compare it to other diffusion-based methods. DiffuSeq \citep{gong2022diffuseq} is a recently proposed text diffusion model using an encoder-only Transformer structure. 
We also compare with concurrently proposed CDCD \citep{dieleman2022continuous} on machine translation.

\subsection{Implementation Details}

We use a 6 layers encoder-decoder Transformer \citep{vaswani2017attention} with GeLU activation \citep{Hendrycks2016BridgingNA}. 
For the diffusion process, we set the maximum diffusion step $T$ to 2000, and use the \textit{sqrt} schedule from DiffusionLM \citep{li2022diffusion} to initialize the adaptive time schedule.  
For translation tasks, we construct vocabulary using BPE \citep{sennrich-etal-2016-neural}. The vocabulary size is set to 10,000 for IWSLT14 and 32,768 for WMT14. For other tasks, we use the vocabulary of \texttt{bert-base-uncased} \citep{devlin-etal-2019-bert}.


For training of SeqDiffuSeq, we use a learning rate of $10^{-4}$ with 10,000 warm-up steps and a linearly-decreasing schedule. The proposed adaptive noise schedule is updated every 20,000 training steps and $K$ is set to 20. We explore maximum Bayes risk (MBR) decoding \citep{koehn-2004-statistical} following previous research \citep{li2022diffusion} for improving generation quality during inference. Details on experiment settings and MBR are in Appendix \ref{app:imple_detail}.

\begin{table*}[t]
    \caption{Ablation studies on IWSLT14, QQP and Wiki-Auto. S-BLEU represents Sacre-BLEU. BERTSco. represents BERTScore. Self-Cond. and Apt. Sche. are short for self-conditioning and adaptive noise schedule. }
    \label{tab:lab-ablation}
    \centering
    \resizebox{1.0\columnwidth}{!}{
    \begin{tabular}{lc|cc|cc|cc|c}
    \hline
     &&\multicolumn{2}{|c|}{{IWSLT14}}&\multicolumn{2}{|c|}{Paraphrase}&\multicolumn{2}{|c|}{Text Simplification}\\
     &&EN-DE&DE-EN&\multicolumn{2}{|c|}{QQP}&\multicolumn{2}{|c|}{Wiki-Auto}&Avg. $\Delta$BLEU\\
     &&S-BLEU&S-BLEU  & BLEU & BERTSco.& BLEU & BERTSco.& \\
     \hline
     $\quad$ SeqDiffuSeq &$\mathcal{A}$&21.96&30.16&23.28&83.91&37.09&82.11&-\\
     $\mathcal{A}$ w/o Apt. Sche.& $\mathcal{B}$&19.89 &28.60 &21.82 & 81.78  &33.04&79.74&-2.29 \\
     $\mathcal{A}$ w/o Self-Cond.&$\mathcal{C}$&20.76&28.28&21.64&81.45&36.46&81.62&-1.34\\
     $\mathcal{C}\ $ w/o Apt. Sche.&$\mathcal{D}$&17.50&24.39&19.73&79.95&28.03&76.06&-5.71 \\
     \hline  
    \end{tabular}
    }
\end{table*}

\begin{figure*}[t]
    \centering
    \resizebox{1.0\columnwidth}{!}{
    \includegraphics{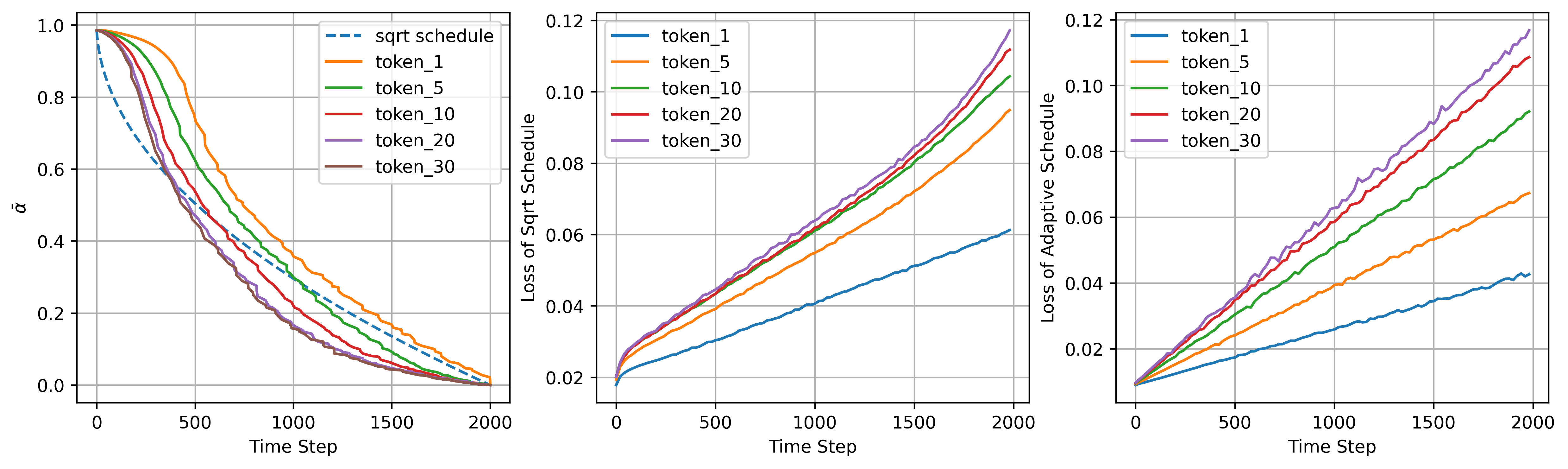}
    }
    \caption{The left figure depicts the adaptive noise schedule at different token positions on IWSLT14 DE-EN dataset. The middle and right figures show the loss for each time step at different token positions with and without adaptive noise schedule, respectively. Best viewed in color.}
    \label{fig:schedule}
\end{figure*}

\subsection{Main Results}


To assess the generation quality of each model, we use BLEU \citep{papineni-etal-2002-bleu} and BERTScore \citep{Zhang2020BERTScore} as metrics. We also use distinct uni-gram (dist.1) to measure the word diversity within generated sentences. A high dist.1 score indicates fewer repeated words. For machine translation tasks, we additionally consider SacreBLEU \citep{post-2018-call}. The results are listed in Table \ref{tab:main-result}. To better present the generation performance, we provide human evaluation results in Appendix \ref{app:human}.

Primarily, for text generation quality, our proposed \name achieves much better performance measured by BLEU than DiffuSeq and other baselines with single generation on QQP, Wiki-Auto, and Quasar-T. On Wiki-Auto and Quasar-T, \name even achieves better performance with single generation than recently proposed DiffuSeq with MBR of 10 candidates. 
When incorporating with MBR, \name enjoys a boost of performance and achieves superior results over all baselines on QQP, Wiki-Auto, and Quasar-T. The performance is better than the pre-trained then fine-tuned GPT-2 with more parameters on Wiki-Auto and QQP. This indicates that \name can generate texts with good quality for sequence-to-sequence tasks (except CCD that all models have inferior performance). 
On translation tasks, the performance lags behind the AR Transformers baseline consistently across different datasets, while compared with NAR methods, \name consistently surpasses 
CMLM with 1 refinement iteration by 6.32 and 6.75 averaged points across four datasets without and with MBR.
CMLM with 4 iterations has better performance.
When comparing with CDCD, the performance with and without MBR are competitive on WMT14 EN-DE while the performance is worse on DE-EN.
For diversity within sequences, texts generated by \name have fewer repeated words averagely than Transformers and DiffuSeq.

The largest improvement over MBR inference with 10 candidates is 1.06 BLEU score on QQP. The amount of this marginal improvement is consistent with concurrently proposed CDCD on WMT14. We will give more in-depth analyses of MBR in the following sections.

\section{Analysis and Discussion}
\subsection{Ablation Study}

To verify the effectiveness of the proposed techniques in SeqDiffuSeq, we conduct ablation studies on QQP, Wiki-Auto, and IWSLT14. As shown in Table \ref{tab:lab-ablation}, after removing the adaptive noise schedule from \name and instead using the fixed \textit{sqrt} schedule proposed in DiffusionLM ($\mathcal{B}$), the performance drops consistently and the BLEU scores decrease by 2.29 on average. Without self-conditioning ($\mathcal{C}$), the performance also degrades by 1.34 on average. By further removing adaptive noise schedule ($\mathcal{D}$), the performance drops sharply by 5.71 on average and the largest drop in terms of BLEU is 8.43 on Wiki-Auto. 
Comparing adaptive noise schedule and self-conditioning technique, it is illustrated that our proposed adaptive noise schedule brings larger improvement and two techniques are complementary to each other. 

\subsection{Time Schedule}

It is verified in the ablation study that the proposed adaptive noise schedule can improve sequence-to-sequence text generation. On the IWSLT14 DE-EN dataset, we visualize the adaptive noise schedules as well as the loss at each time step with and without adaptive noise schedule. For the adaptive noise schedule, we plot $\bar{\alpha}_t^i$ at different token positions $i$ against the diffusion time step $t$. And for losses, we plot averaged training losses $\mathcal{L}^i_t$ at each position $i$ against time step $t$.
Depicted in Figure \ref{fig:schedule}, the dashed line in the first sub-figure shows the \textit{sqrt} schedule, while the other lines represent the noise schedules at different token positions. The figure shows that the adaptive noise schedules deviate from the \textit{sqrt} schedule. At both ends of time steps, the adaptive noise schedules are flatter compared to \textit{sqrt} schedule, especially for tokens at larger position orders.
Besides, adaptive noise schedules are diverse for different positions, although the trends along time steps are similar. For the token positions at larger orders, the noise schedule lines move toward the lower-left direction. Therefore, at each time step, the tokens at earlier positions have smaller noise than later positions. The information of tokens on the left is recovered earlier at each step. \name resembles the left-to-right generation of texts. Through a case study in Appendix \ref{app:case}, the phenomenon is also verified.

\begin{wraptable}[7]{r}{0.48\textwidth}
    \small
     \caption{Inference time on QQP on one NVIDIA V100 GPU. The inference batch size is set to 50 and the overall time step is set to 2000 for both models. }
     \label{tab:inference speed}
    \centering
    \begin{tabular}{l|cc}
         &Time &Accelerate\\
         \hline
         DiffuSeq&317 sec.&-\\
         SeqDiffuSeq&89 sec.& $\times$3.56\\
    \end{tabular}
\end{wraptable}

Comparing the second and third sub-figures, the losses $\mathcal{L}^i$ with adaptive noise schedule increase linearly with respect to time steps as expected. At each time step, the losses at earlier token positions are smaller, indicating earlier tokens are easier to generate for \name. More visualizations on other datasets are listed in Appendix \ref{app:more_schedule}.

\begin{wrapfigure}[34]{r}{0.48\textwidth}
    \centering
    \small
    \includegraphics[width=1.0\linewidth]{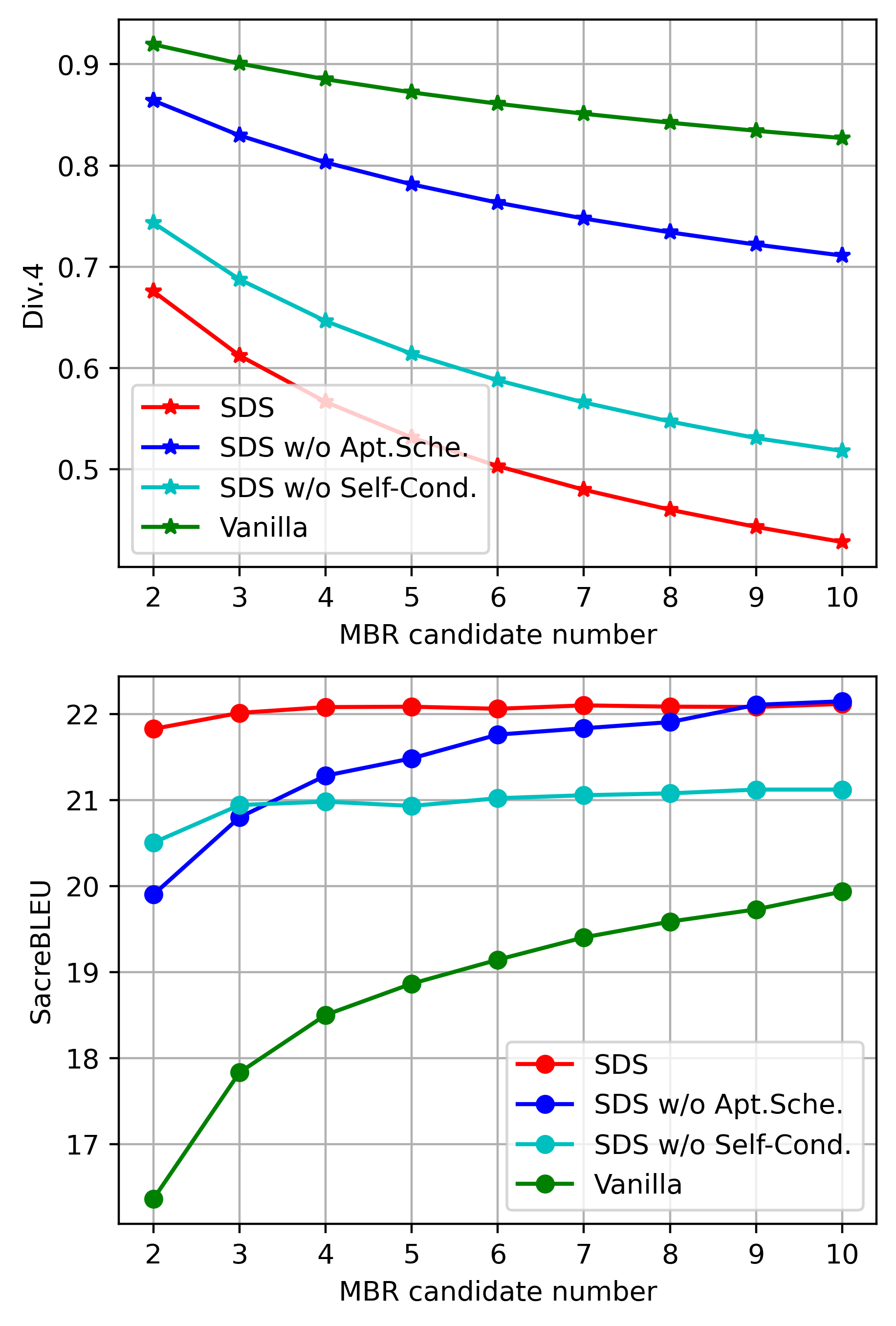}
    \caption{The top figure plots the sequence-level Div.4 score against different MBR candidate numbers on IWSLT14 EN-DE. The bottom figure plots SacreBLEU against different MBR candidate numbers. SDS represents SeqDiffuSeq. Best viewed in color.}
    \label{fig:bleu-mbr}
\end{wrapfigure}

\subsection{Inference Speed}

We compare SeqDiffuSeq with DiffuSeq in terms of inference time in Table \ref{tab:inference speed}. Our SeqDiffuSeq achieves 3.56 times acceleration generating one batch of text samples. The acceleration mainly originated from: (1) SeqDiffuSeq only requires forward computation of encoder once, while DiffuSeq needs to run forward computation for the input sequences for each diffusion step; 
(2) At each time step, SeqDiffuSeq only models the output sequence, while DiffuSeq has to model the concatenation of both input and output sequences. 

\subsection{MBR Inference}

It is shown in Table \ref{tab:main-result} that MBR with 10 candidates improves DiffuSeq to more than 6 BLEU score, while improves \name by 1.06 BLEU score on QQP. In Figure~\ref{fig:bleu-mbr}, we plot SacreBLEU scores and Diverse 4-gram (Div.4) scores \citep{Deshpande2018FastDA} against MBR candidate numbers. Div.4 measures the proportion of distinct 4-grams in a set of generated sequences. A higher Div.4 score means better sequence-level diversity by different generation runs. The figure shows that the self-conditioning technique and adaptive noise schedule make the text diffusion model generate less diverse sequences, and the single generated sequence will have higher quality with both techniques. Self-conditioning technique and adaptive noise schedule deliver a trade-off between generation quality and generation diversity. With both techniques, MBR inference is needless to generate high-quality samples for \name resulting in a more efficient generation procedure. We also propose a new sampling scheme to compensate the marginal MBR improvements for \name which is discussed in detail in Appendix \ref{app:qsample}.

\section{Conclusion}

In this work, we explore to approach sequence-to-sequence text generation with continuous diffusion models. We propose SeqDiffuSeq which uses an encoder-decoder Transformers architecture to learn the denoising function. In order to improve text generation performance, the denoising function in SeqDiffuSeq is integrated with self-conditioning technique. SeqDiffuSeq also includes a newly proposed adaptive noise schedule which makes the denoising difficulty evenly distributed across all time steps and assigns exclusive noise schedules for tokens from different positional orders. Through experiments, we illustrate the superior performance of SeqDiffuSeq in terms of generation quality and inference speed and provide insights into our proposed adaptive noise schedule technique.

\bibliography{custom,anthology}
\bibliographystyle{achemso}

\newpage
\appendix

\section{Limitation}

Diffusion models generate high-quality synthetic samples through thousands of iterations in the reverse process. Thousands of reverse process iterations require a huge amount of forward propagation computation of Transformers model which is computationally costly, although we save nearly four times of computational budget for one forward computation compared to the previous diffusion-based model DiffuSeq. In the domain of vision synthetic, there exists research to profoundly reduce the time step needed for generation \citep{ddim}. Reducing the reverse steps for text generation would be a promising direction for future research.

As shown in the discussion, equipping text diffusion models with self-conditioning and adaptive noise schedules can profoundly increase the generation quality. However, such quality improvement is at the cost of generation diversity under different random seeds. This leads to marginal MBR inference improvements. Although we propose a compensation discussed in Appendix \ref{app:qsample}. The in-depth discussion on improving \name generation diversity is left to future research.

\section{Ethic Statements and Boarder Impact}

The datasets and baseline models used in our research are publicly available. 
Diffusion models, previously successful in vision, face challenges in NLP due to discrete token sequences. Promising results have been shown in DiffusionLM \citep{li2022diffusion} and DiffuSeq \citep{gong2022diffuseq}, but both works use encoder-only models and have limitations in scalability and efficiency.
This research explores and improves the diffusion-based sequence-to-sequence text generation models. 
Our work alters to encoder-decoder Transformers which are widely applied in recent LLMs such as FLAN-T5 \citep{chung2022scaling} for better scalability, potential, and sampling speed acceleration (Section 6.3). 
Our work also incorporates novel techniques like self-conditioning and adaptive noise schedules, outperforming several AR and NAR baselines. \name demonstrates the feasibility of encoder-decoder diffusion models for sequence-to-sequence tasks and may serve as a starting point for future exploration of text diffusion models' potential, serving as another method approaching sequence-to-sequence text generation besides widely implemented AR and NAR models. 
Considering the excellent performance of diffusion models in other domains such as vision, text diffusion models have great potential in generating text sequences with high quality and may be an emerging framework of text generation.

\section{Derivation of Posterior}
\label{app:derive_post}
Given $z_t\sim q(z_t|z_{t-1}) = \mathcal{N}(z_t;\sqrt{\alpha_t}z_{t-1}, (1-\alpha_t)I)$, we can reparameterize $z_t=\sqrt{\alpha_t}z_{t-1}+\sqrt{1-\alpha_t}\epsilon_t$. Then, recursively, 
\begin{align}
    z_t& =\sqrt{\alpha_t}(\sqrt{\alpha_{t-1}}z_{t-2}+\sqrt{1-\alpha_{t-1}}\epsilon_{t-1})+\sqrt{1-\alpha_t}\epsilon_t \nonumber\\
    &=\sqrt{\bar{\alpha}_t}z_0+\sqrt{1-\bar{\alpha}_t}\epsilon_t \nonumber\\
    &\sim \mathcal{N}(z_t;\sqrt{\bar{\alpha}_t}z_0, (1-\bar{\alpha}_t)I). \label{appequ:prior_z0}
\end{align}
Therefore, $q(z_t|z_0)=\mathcal{N}(z_t;\sqrt{\bar{\alpha}_t}z_0, (1-\bar{\alpha}_t)I)$. According to Bayes rule, we have:
\begin{align}
    q(z_{t-1}|z_t,z_0) = \frac{q(z_{t}|z_{t-1})q(z_{t-1}|z_0)}{q(z_t|z_0)},
\end{align}
since $q(z_{t}|z_{t-1})$ and $q(z_{t-1}|z_0)$ are all Gaussian distributed, we will have:
\begin{align}
    q(z_{t-1}|z_t,z_0)=\mathcal{N}(z_{t-1};\tilde{\mu}(z_0,z_{t}), \tilde{\beta}_{t}I),
\end{align}
where
\begin{align}
    \tilde{\mu}(z_0,z_{t})=\frac{\sqrt{\bar{\alpha}_{t-1}}\beta_{t}}{1-\bar{\alpha}_{t}}{z_0}+\frac{\sqrt{\alpha_{t}}(1-\bar{\alpha}_{t-1})}{1-\bar{\alpha}_{t}}z_{t}, \\
\bar{\alpha}_t=\prod_{s=1}^t\alpha_s, \quad \beta_t=1-\alpha_t, \quad \tilde{\beta}_{t} = \frac{1-\bar{\alpha}_{t-1}}{1-\bar{\alpha}_t}\beta_t.
\end{align}

\section{Derivation of Training Objective}
\label{app:derive_obj}

We present the detailed derivation of training objective following \citet{ho2020denoising,li2022diffusion}. As mentioned in main texts, the forward process successively perturbs the real-world sample $z_0$ with random noise, where $z_0$ gradually changes to $z_T$ for a $T$-time step diffusion process. $z_T$ can be approximately regarded as pure random noise which follows standard Gaussian distribution in our case. We define the forward process as follows:
\begin{align}
    q(z_t|z_{t-1}) = \mathcal{N}(z_t;\sqrt{\alpha_t}z_{t-1}, (1-\alpha_t)I),
\end{align}
where $\alpha_t$ controls the noise level at each time step $t$.

For the reverse process, we learn a parameterized denoising distribution $p_\theta(z_{t-1}|z_t, w_x, t)$. By successively sampling from $p_\theta$, a synthetic real-world sample $z_0$ can be recovered from pure random noise $z_T$. 

The training objective of diffusion model is to minimize the negative likelihood of data distribtuion, which is:
\begin{align}
    \tilde{\mathcal{L}} = \mathbb{E}[-\log p_\theta(z_0)],
\end{align}
then with the forward and revser process defined as above, we can derive the variational bound for the objective $\tilde{\mathcal{L}}$:
\begin{align}
    \tilde{\mathcal{L}}= &\mathbb{E}_{q(z_0)}[-\log p_\theta(z_0)] \nonumber \\
    \le &\mathbb{E}_{q(z_{0:T})}\left[-\log \frac{p_\theta(z_{0:T})}{q(z_{1:T}|z_0)}\right] \nonumber \\
    =&\mathbb{E}_{q(z_{0:T})}\left[-\log p(z_T)-\sum_{t\ge 1}\log\frac{p_\theta(z_{t-1}|z_t)}{q(z_t|z_{t-1})}\right] \nonumber \\
    =&\mathbb{E}_{q(z_{0:T})}\left[-\log p(z_T)-\sum_{t> 1}\log\frac{p_\theta(z_{t-1}|z_t)}{q(z_t|z_{t-1})} -\log\frac{p_\theta(z_0|z_1)}{q(z_1|z_0)}\right]. 
    \label{appequ:ori_loss}
\end{align}

In our sequence-to-sequence settings, following the notations in Section \ref{approach}, we let the denoising distribution $p_\theta$ condition on the input sequence $w_x$, which is $p_\theta(z_{t-1}|z_t, w_x)$. Besides, with the Markov transition extensions of embedding mapping transition $q_\phi(z_0|w_y)$ in the forward process and rounding transition $\tilde{p}_\phi(w_y|z_0)$ in the reverse process, the objective in Equation \ref{appequ:ori_loss} can be extended as:
\begin{align}
    \mathcal{L}=& \mathbb{E}_{q_\phi(z_{0:T},w_x,w_y)}\bigg[-\log p(z_T)-\sum_{t> 1}\log\frac{p_\theta(z_{t-1}|z_t, w_x)}{q(z_t|z_{t-1})} \nonumber\\
    &-\log\frac{p_\theta(z_0|z_1, w_x)}{q(z_1|z_0)}-\log\tilde{p}_\phi(w_y|z_0) + \log q_\phi(z_0|w_y)\bigg]. \label{appequ:ori_loss_s2s}
\end{align}

By Bayes rule, we can derive the posterior distribution of $q$ with respect to $z_{t-1}$:
\begin{align}
    q(z_{t-1}|z_t, z_0) &= \frac{q(z_t|z_{t-1}, z_0)q(z_{t-1}|z_0)}{q(z_t|z_0)} ,
\end{align}
then, we have:
\begin{align}
    q(z_t|z_{t-1}) & = \frac{q(z_{t-1}|z_t, z_0)q(z_t|z_0)}{q(z_{t-1}|z_0)} \label{appequ:q_post}.
\end{align}
We substitute $q(z_t|z_{t-1}), \forall t>1$ in Equation \ref{appequ:ori_loss_s2s} with Equation \ref{appequ:q_post}:
\begin{align}
    {\mathcal{L}}_{VB} =&\mathbb{E}_{q_\phi}\bigg[-\log\frac{p(z_T)}{q(z_T|z_0)}-\sum_{t> 1}\log\frac{p_\theta(z_{t-1}|z_t, w_x)}{q(z_{t-1}|z_t, z_0)}  \nonumber\\
    &-\log{p_\theta(z_0|z_1, w_x)}-\log\tilde{p}_\phi(w_y|z_0) + \log q_\phi(z_0|w_y)\bigg] \label{appequ:lvb}
\end{align}
For the time step $t,t>1$, the terms $-\mathbb{E}_{q_\phi}\left[\log\frac{p_\theta(z_{t-1}|z_t)}{q(z_{t-1}|z_t, z_0)}\right]$ between two Gaussian
distributions has a closed form solution, following \citet{li2022diffusion,ho2020denoising}, we have:
\begin{align}
    &-\mathbb{E}_{q_\phi}\left[\log\frac{p_\theta(z_{t-1}|z_t)}{q(z_{t-1}|z_t, z_0)}\right] \nonumber \\
    =& \mathbb{E}_{q_\phi}\left[\left\|\frac{1}{2\sigma^2_t}\left(\tilde{\mu}_\theta(z_{t}, w_x, t)-\tilde{\mu}(z_0,z_t)\right)\right\|^2 \right]+ C \nonumber \\
    \propto&\mathbb{E}_{q_\phi}\left[\left\|\tilde{\mu}_\theta(z_{t}, w_x, t)-\tilde{\mu}(z_0,z_t)\right\|^2\right],\label{appequ:mu0_mse}
\end{align}
where $C$ is a constant and $\sigma^2_t=\tilde{\beta}_t$, then substituting $\tilde{\mu}$ and $\tilde{\mu}_\theta$ by Equation \ref{equ:posterior_mean} and \ref{equ:denoise_mean}, we have:
\begin{align}
    &\left\|\tilde{\mu}_\theta(z_{t}, w_x, t)-\tilde{\mu}(z_0,z_t)\right\|^2 \nonumber \\
    =& \frac{\sqrt{\bar{\alpha}_{t-1}}\beta_{t}}{1-\bar{\alpha}_{t}}\left\|z^0_\theta(z_{t}, w_x, t)-z_0\right\|^2 \nonumber \\
    \propto&\left\|z^0_\theta(z_{t}, w_x, t)-z_0\right\|^2. \label{appequ:z0_mse}
\end{align}
After omitting $\frac{1}{2\sigma^2_t}$ and $\frac{\sqrt{\bar{\alpha}_{t-1}}\beta_{t}}{1-\bar{\alpha}_{t}}$ for any $t>2$, and substituting terms $-\mathbb{E}_{q_\phi}\left[\log\frac{p_\theta(z_{t-1}|z_t)}{q(z_{t-1}|z_t, z_0)}\right]$ in Equation \ref{appequ:lvb} with Equation \ref{appequ:mu0_mse}, \ref{appequ:z0_mse}, we have the simplified loss function:
\begin{align}
\tilde{\mathcal{L}}_{simple}=&\mathbb{E}_{q_\phi}\bigg[-\log\frac{p(z_T)}{q(z_T|z_0)}+\sum_{t> 1}\left\|z^0_\theta(z_{t}, w_x, t)-z_0\right\|^2 \nonumber\\
    &-\log\frac{{p_\theta(z_0|z_1, w_x)}}{ q_\phi(z_0|w_y)}-\log\tilde{p}_\phi(w_y|z_0) \bigg].
\end{align}
We can further substituting terms $-\mathbb{E}_{q_\phi}\left[\log\frac{p(z_T)}{q(z_T|z_0)}\right]$ and $-\mathbb{E}_{q_\phi}\left[\log\frac{{p_\theta(z_0|z_1, w_x)}}{ q_\phi(z_0|w_y)}\right]$ similarly with:
\begin{align}
    &-\mathbb{E}_{q_\phi}\left[\log\frac{p(z_T)}{q(z_T|z_0)}\right]\propto \mathbb{E}_{q_\phi}\left[\|\mu_T(z_0)\|^2\right],\label{appequ:mse_zT}\\
    &-\mathbb{E}_{q_\phi}\left[\log\frac{{p_\theta(z_0|z_1, w_x)}}{ q_\phi(z_0|w_y)}\right]\propto\mathbb{E}_{q_\phi}\left[\|z^0_\theta(z_{1}, w_x, 1)-g_\phi(w_y)\|\right]^2, \label{appequ:mse_round}
\end{align}

where $\mu_T(z_0)=\sqrt{\bar{\alpha}_T}z_0$.Therefore we can derive $\mathcal{L}_{simple}$ in Equation \ref{equ:simple} by subsitituting terms in $\tilde{\mathcal{L}}_{simple}$ with Equation \ref{appequ:mse_zT} and \ref{appequ:mse_round}:

\begin{align}
    \mathcal{L}_{simple}&=\mathbb{E}_{q_\phi}\bigg[\sum_{t>1} \|z^0_\theta(z_{t}, w_x, t)-z_0\|^2 \nonumber\\
    &\quad+\|\mu_T(z_0)\|^2+\|z^0_\theta(z_{1}, w_x, 1)-g_\phi(w_y)\|^2 \quad-\log{\tilde{p}_\phi(w_y|z_0)}\bigg]\\
    &=\mathbb{E}_{q_\phi(z_0,w_x,w_y)}\bigg[\sum_{t=2}^T\mathbb{E}_{q(z_t|z_0)} \|z^0_\theta(z_{t}, w_x, t)-z_0\|^2 \nonumber\\
    &\quad+ \|\mu_T(z_0)\|^2+\|z^0_\theta(z_{1}, w_x, 1)-g_\phi(w_y)\|^2 \quad-\log{\tilde{p}_\phi(w_y|z_0)}\bigg].
\end{align}

\section{Datasets}
\label{app:data_stat}

We conduct experiments on following datasets. The data statistics and licenses are shown in Table~\ref{tab:data_stat} and~\ref{tab:data_lisen}. 

\noindent \textbf{Quora Question Pairs} (QQP) \citep{quora-question-pairs} is a paraphrase identification dataset. We use the positive pairs as the paraphrase generation task. The models need to generate a restatement expressing the same meaning to the given sentence. \\
\textbf{Wiki-Auto} \citep{jiang2020neural} is a text simplification dataset to revise a complex text with simplified grammar and word choices. The dataset aligns sentences between English Wikipedia and Simple English Wikipedia with automatic pre-processing and identifying procedure. \\
\textbf{Quasar-T} \citep{dhingra2017quasar} is a question-answering dataset containing trivia questions paired with answers and contexts.
We use the dataset for evaluating question generation which aims to generate related questions with given contexts. We use the pre-processed data from \citet{lin-etal-2018-denoising} following \citet{gong2022diffuseq}. \\
\textbf{Commonsense Conversation Dataset} (CCD) \citep{zhou2018commonsense} is extracted from single-round dialogues on Reddit and is used for evaluating open domain dialogue generation. The task requires generating feedback with commensense knowledge given the dialogue contexts.\\
\textbf{IWSLT14} and \textbf{WMT14} are both widely used benchmarks for machine translation. We use the German(DE)-English(EN) pairs for both directions of translation. We follow fairseq \citep{ott-etal-2019-fairseq} for data pre-processing using Moses script \citep{koehn-etal-2007-moses} and tokenizing the sentences with byte-pair encoding (BPE) \citep{sennrich-etal-2016-neural}. 

\begin{table}[t]
    \caption{The data splits statistics.}
    \label{tab:data_stat}
    \centering
    \begin{tabular}{l|ccc}
        Dataset & Train size & Dev size & Test Size \\
        \hline
        QQP &  144,715 & 2,048 & 2,500\\
        Quasar-T & 116,953 & 2,048 & 10,000 \\
        Wiki-Auto &677,751&2,048&5,000\\
        CCD & 3,382,137& 2,048 & 10,000\\
        IWSLT14 &160,239&7,283&6,750\\
        WMT14 &4,475,414&45,206&3,003\\
    \end{tabular}
\end{table}

\begin{table}[t]
    \caption{The license of data used in experiments.}
    \label{tab:data_lisen}
    \centering
    \begin{tabular}{l|c}
        QQP & CC-BY-SA-3.0 from GLUE \\
        Quasar-T &  BSD-2-Clause license\\
        Wiki-Auto & Unspecified, Wikipedia by CC-BY-SA-3.0\\
        CCD & Apache License 2.0\\
        IWSLT14 &CC-BY-NC-ND-4.0\\
        WMT14 & Unspecified\\
    \end{tabular}
\end{table}

\section{Implementation Details}
\label{app:imple_detail}

\subsection{Details on Experiment Setting}
Here we give details for the implementation details of our experiments. For the Transformers structure and model training, we list detailed design in Table \ref{tab:transformer}. For all the tasks, the set the maximum training step to 1000,000 and save checkpoints every 10,000 steps. We select the best checkpoint on the development set. For WMT14 task, we use batch size 1024 while for other tasks we use batch size 128. 
For training on each datasets, we train for one run on NVIDIA A100 GPUs with 80GB memory.
For inference, we set the maximum time step to $T=2000$, and we do not use the clamping trick as proposed in DiffusionLM \citep{li2022diffusion}, since the clamping trick does not consistently improve the generation quality across datasets. 

\begin{table}[t]
    \caption{Translation represents the machine translation tasks on IWLST14 and WMT14. Non-Translation represents the Paraphase, Text Simplification, Queation Generation and Dialogue tasks on QQP, Wiki-Auto, Quasar-T and CCD respectively.}
    \label{tab:transformer}
    \centering
    \begin{tabular}{l|ccc}
        Tasks& Translation & Non-Translation \\
        \hline
        Encoder Layer &  6 & 6\\
        Decoder Layer & 6 & 6 \\
        Head Number & 8& 12\\
        Hidden Dimension&512&768\\
        FFN Dimension &2048&3072\\
        Embedding Dimension &128 &128\\
        Max. Input Length &128&128\\
        Max. Output Length&64&64 \\
        Dropout & 0.3&0.1 \\
    \end{tabular}
    
\end{table}

\subsection{Details on MBR}

Following DiffusionLM \citep{li2022diffusion}, we apply Minimum Bayes Risk (MBR) decoding for one single generation output with improved quality.
For each sample, MBR decoding uses a generated sequences candidate set $\mathcal{C}$ and finds the candidate sequence $s^*$ that minimize a expected risk $R$:
\begin{align}
    s^* = \arg\min_{s\in \mathcal{C}} R(s) = \arg\min_{s\in \mathcal{C}} \frac{1}{|\mathcal{C}|}\sum_{s'\in\mathcal{C}}r(s,s'),
\end{align}
where $r(\cdot, \cdot)$ is a specific risk function and we use the negative BLEU score following DiffusionLM and sequence candidates in the candidate set $\mathcal{C}$ are generated from the diffusion models under different random seeds.

\section{Sampling by Prior}
\label{app:qsample}
\begin{figure*}[t]
    \centering
    \resizebox{1.0\columnwidth}{!}{
    \includegraphics{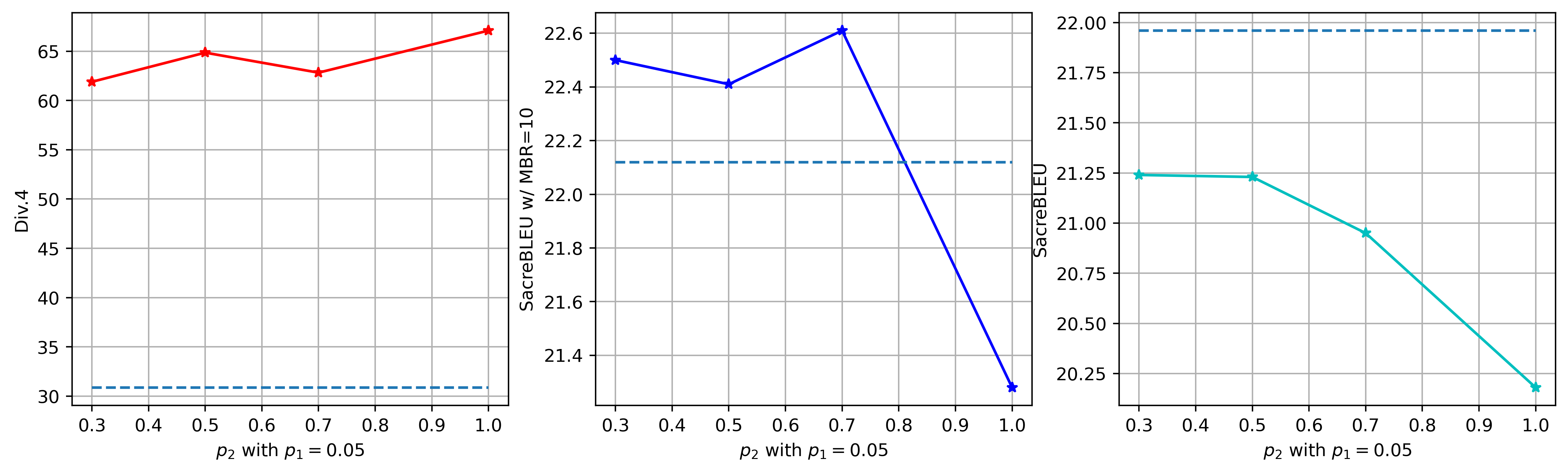}
    }
    \caption{The figures from left to right plot the diversity, SacreBLUE with MBR=10 and SacreBLEU for single candidates against $p_2$ on IWSLT14 EN-DE dataset with $p_1=0.05$ fixed, repectively. The dashed lines in each figure represents the default generation results of SeqDiffuSeq.}
    \label{appfig:qsample1}
\end{figure*}
\begin{figure*}[t]
    \centering
    \resizebox{1.0\columnwidth}{!}{
    \includegraphics{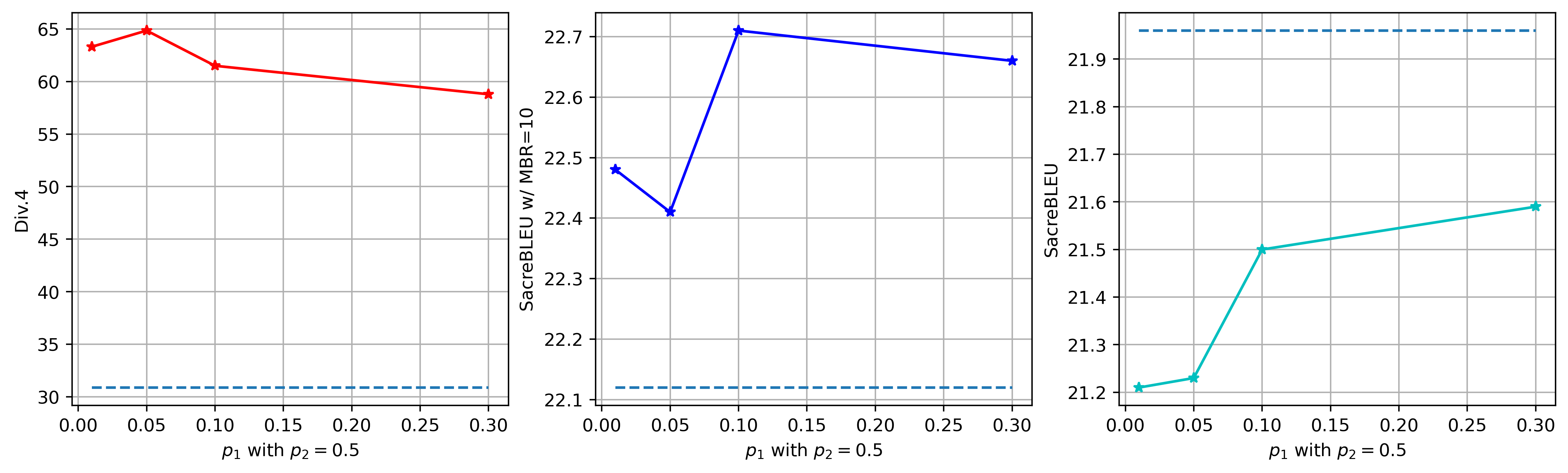}
    }
    \caption{The figures from left to right plot the diversity, SacreBLUE with MBR=10 and SacreBLEU for single candidates against $p_1$ on IWSLT14 EN-DE dataset with $p_2=0.5$ fixed, repectively. The dashed lines in each figure represents the default generation results of SeqDiffuSeq.}
    \label{appfig:qsample2}
\end{figure*}
Since at each time step $t$, the Transformers denoising function $z_\theta^0$ models the prediction $\hat{z}_0^t$ of target output sequences. In the reverse process, sampling $z_{t-1}$ is according to the denoising distribution $p_\theta$ as:
\begin{align}
    p_\theta(z_{t-1}|z_{t}, w_x) = \mathcal{N}(z_{t-1}; \tilde{\mu}_\theta(z_{t}, w_x, t), \tilde{\beta}_{t}I). \label{equ:denoising_dist}
\end{align}
However, we can also use the prior distribution $q$ in the forward process to generate $z_{t-1}$, which is:
\begin{align}
    z_{t-1}\sim &q(z_{t-1}|\hat{z}_0^t) \nonumber \\
    =&\mathcal{N}(z_{t-1};\sqrt{\bar{\alpha}_{t-1}}\hat{z}_0^t, (1-\bar{\alpha}_{t-1})I). \label{appequ:qsample}
\end{align}
Comparing to generation by Equation \ref{equ:denoising_dist}, using Equation \ref{appequ:qsample} theoretically have larger variance. 
\begin{align}
    1-\bar{\alpha}_{t-1} \ge\tilde{\beta}_{t} = \frac{1-\bar{\alpha}_{t-1}}{1-\bar{\alpha}_t}\beta_t,
\end{align}
because
$\frac{\beta_t}{1-\bar{\alpha}_t}=\frac{1-\alpha_t}{1-\bar{\alpha}_t} \le 1$ where $\alpha_t<1, \forall t$ and $\bar{\alpha}_t=\prod_{s=1}^t\alpha_s$.

To increase the sequence level diversity, we experiment with randomly replacing the denoising distribution $p_\theta$ by high variance distribution in Equation \ref{appequ:qsample} in the reverse process during generation. We denote the replacing probability as $p_1$.

Besides, considering the variance difference between the two sampling distribution are larger at earlier time step in the reverse process, we also explore to only replace the sampling distribution in the first $p_2$ percent of time steps. We generate 10 candidate output sentences for each sample under different random seeds to compute Div.4 and SacreBLEU scores.

As shown in the left subfigure of Figure \ref{appfig:qsample1}, when fixing the replacing probability to 0.05, the generation diversity are consistently and profoundly improved. In the right subfigure, the generation quality consistently degrades when replacing the denoising distribution when generation, even though the replacing probability is low. In the middle subfigure, we can see that although the generation quality degrades for each candidate, the final output sequences by MBR may improve with proper $p_2$. In Figure \ref{appfig:qsample2}, we can get similar results when fixing $p_2=0.5$. In the middle subfigure of Figure \ref{appfig:qsample2}, the final output sequences are consistently better with different $p_1$.

To conclude, it is shown that replacing the sampling distribution from the denoising distribution $p_\theta$ to the prior distribution $q$ can provide a trade-off between the generation diversity and generation quality. With a proper combination of $p_1$ and $p_2$, the generation quality of \name with the aid of MBR can be further improved. The benefits of sampling with the prior distribution $q$ are always neglected in previous research.

\section{More Results on Adaptive Noise Schedule}
\label{app:more_schedule}

\begin{figure*}[t]
    \centering
    \resizebox{1.0\columnwidth}{!}{
    \includegraphics{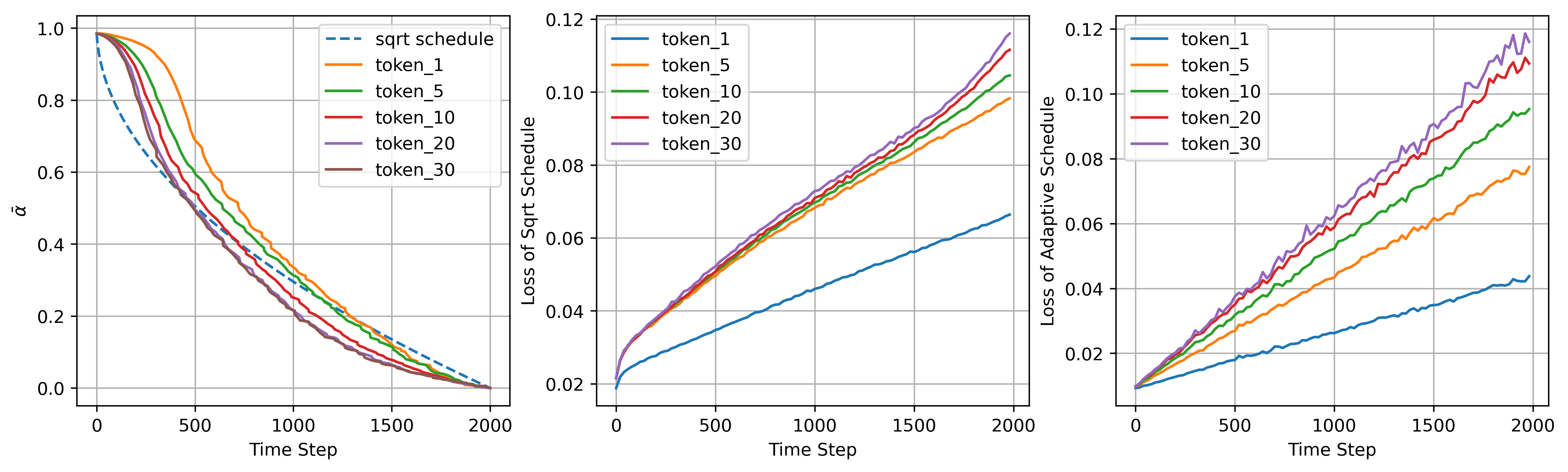}
    }
    \caption{The left figure depicts the adaptive noise schedule at different token positions on IWSLT14 EN-DE dataset. The middle figure shows the loss for each time step at different token positions without the adaptive noise schedule. The right figure shows the loss for each time step at different token positions with the adaptive noise schedule. Best viewed in color.}
    \label{appfig:schedule_ende}
\end{figure*}

\begin{figure*}[t]
    \centering
    \resizebox{1.0\columnwidth}{!}{
    \includegraphics{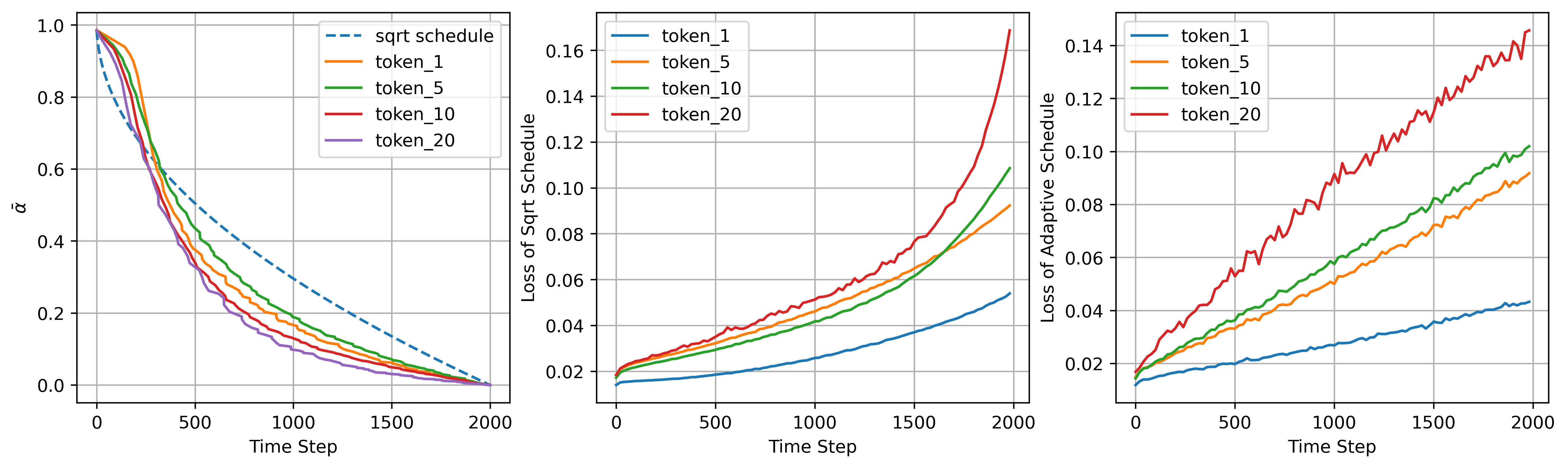}
    }
    \caption{The left figure depicts the adaptive noise schedule at different token positions on QQP dataset. The middle figure shows the loss for each time step at different token positions without the adaptive noise schedule. The right figure shows the loss for each time step at different token positions with the adaptive noise schedule. Best viewed in color.}
    \label{appfig:schedule_qqp}
\end{figure*}

\begin{figure*}[t]
    \centering
    \resizebox{1.0\columnwidth}{!}{
    \includegraphics{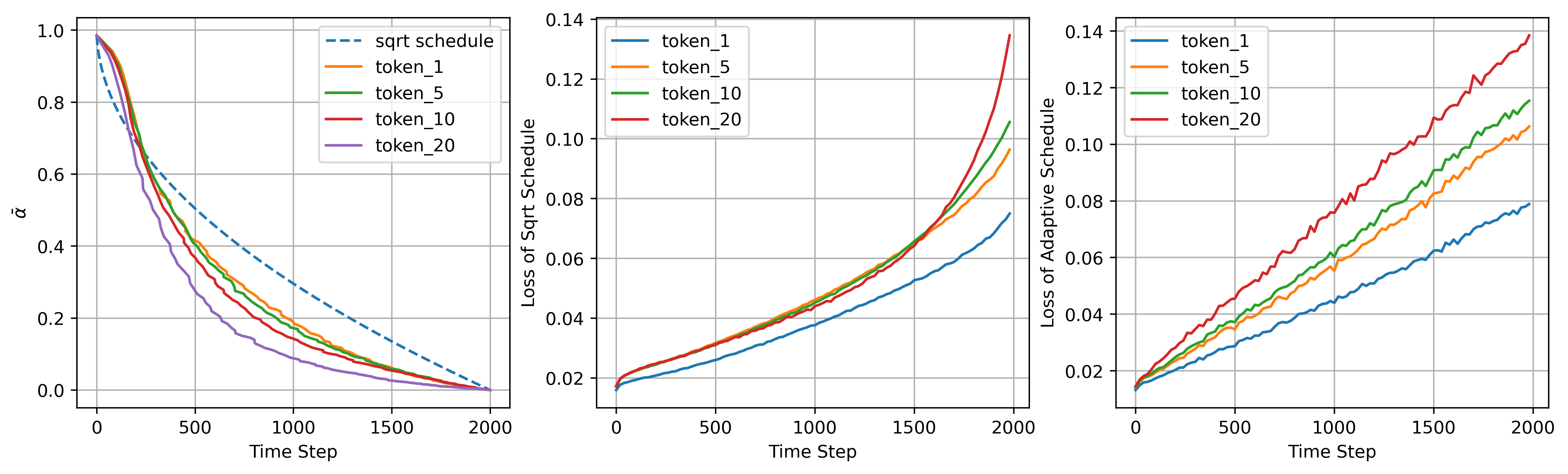}
    }
    \caption{The left figure depicts the adaptive noise schedule at different token positions on Wiki-Auto dataset. The middle figure shows the loss for each time step at different token positions without the adaptive noise schedule. The right figure shows the loss for each time step at different token positions with the adaptive noise schedule. Best viewed in color.}
    \label{appfig:schedule_wiki}
\end{figure*}

We present more visualizations of the learned adaptive noise schedules and the losses for each time step on other datasets. Figure \ref{appfig:schedule_ende}, \ref{appfig:schedule_qqp} and \ref{appfig:schedule_wiki} present the visualizations on IWSLT14 EN-DE, QQP, and Wiki-Auto respectively with the same arrangement as Figure \ref{fig:schedule}. The results from the figures are consistent with those discussed in the main texts. 


\section{Human Evaluation}
\label{app:human}

To better demonstrate the performance of the proposed \name, we conduct human evaluations to compare the generated results of \name to those of DiffuSeq on the paraphrasing task QQP dataset. We randomly sample 100 data points in the test sets and let annotators decide for the same input sequence, which generated text sequence is better, worse, or of similar quality. We compare \name with the previous state-of-the-art text diffusion model DiffuSeq. For fairness, the human evaluations are designed to be blind evaluations (i.e., the annotators are unaware of which model the output sequence is related to).

\begin{figure*}[t]
    \centering
    \resizebox{1.0\columnwidth}{!}{
    \includegraphics{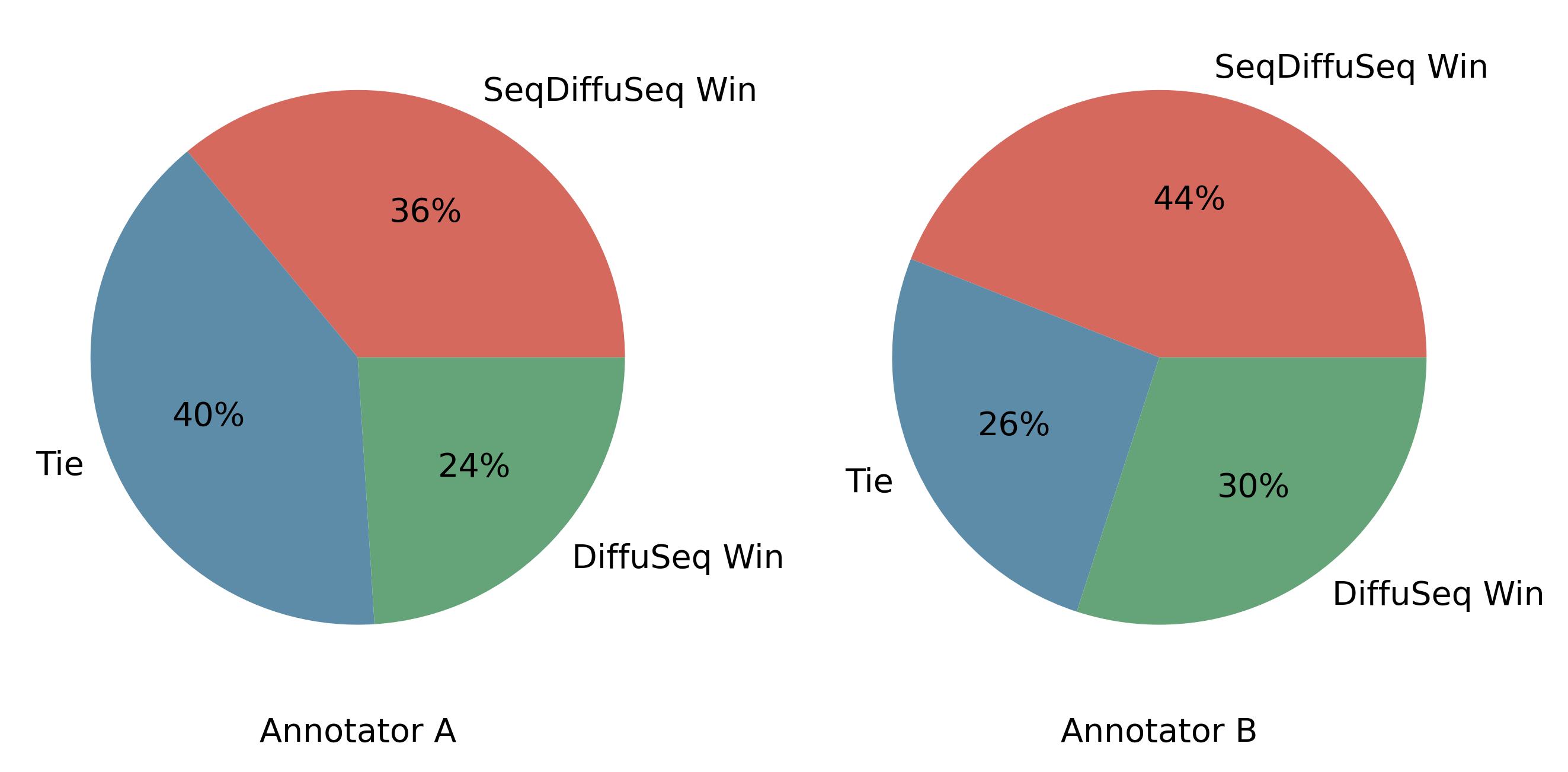}
    }
    \caption{Pie plots of human evaluation results by two different annotators.}
    \label{appfig:human_eval}
\end{figure*}

The human annotators are graduate university students who are proficient in English and are asked to compare the generated sequences based on the following instruction. \textit{Decide which generated output sequence is better based on whether the one is more consistent with the input question, whether the one has higher grammatical and syntactic quality.} 
Figure \ref{appfig:human_eval} shows the human evaluation results.

The results show that both annotators prefer the generated output sequences by \name more. Generated output sequences on QQP from \name win by 36\% and 44\% from two annotators, while those from DiffuSeq only win by 24\% and 30\% respectively. Human evaluation results show that \name can generate text sequences of higher quality than DiffuSeq.

\section{Case Study}
\label{app:case}
\begin{table*}[t]
    \caption{Three cases from QQP. We truncate the selected samples to the first 15 tokens. Generally, SeqDiffuSeq can easily learn to generate [PAD] tokens after the ending token [SEP].}
    \label{tab:casestudy}
    \centering
    \resizebox{1.0\textwidth}{!}{
    \begin{tabular}{c|l}
         Time Step $T-t$ & $z^t$\\
         \hline
         Input Text & How do I read and find my YouTube comments? \\
         400&  [CLS] how do i read in??? [SEP] [PAD] [PAD] [PAD] [PAD] [PAD] \\
         800& [CLS] how do i read my a the? [SEP] [PAD] [PAD] [PAD] [PAD] [PAD] \\
         1200&[CLS] how do i read my youtube comments? [SEP] [PAD] [PAD] [PAD] [PAD] [PAD] \\
         1600& [CLS] how do i read my youtube comments? [SEP] [PAD] [PAD] [PAD] [PAD] [PAD] \\
         2000&[CLS] how do i read my youtube comments? [SEP] [PAD] [PAD] [PAD] [PAD] [PAD]\\
         \hline
         Input Text & How do I use Twitter as a business source? \\
         400& [CLS] how can i use??? a??? [SEP] [PAD] [PAD] \\
         800&[CLS] how can i use?? as a business?? [SEP] [PAD] [PAD] \\
         1200&[CLS] how can i use? twitter as a business source? [SEP] [PAD] [PAD] \\
         1600& [CLS] how can i use? twitter as a business source? [SEP] [PAD] [PAD] \\
         2000&[CLS] how can i use \textcolor{red}{twitter twitter} as a business source? [SEP] [PAD] [PAD] \\
         \hline
         Input Text &What is the funniest joke you know?\\
         400&[CLS] what is the the tot the you? a? [PAD] [PAD] [PAD] \\ 
         800&[CLS] what is the fun?t joke you'for? in? [SEP] \\ 
         1200&[CLS] what is the funniest joke you've ever know? [SEP] \\ 
         1600&[CLS] what is the funniest joke you've ever know? [SEP] \\ 
         2000&[CLS] what is the funniest joke you've ever know? [SEP] \\
    \end{tabular}
    }
    
\end{table*}

We select three illustrative cases and investigate the generation process of SeqDiffuSeq. From the cases, it shows that SeqDiffuSeq can generate reasonable text sequences. The generation process reveals that 

1. SeqDiffuSeq decides the output sequence length by generating [SEP] tokens at the early stage of sampling; 

2. The generation process seems to follow a left-to-right refining order; 

3. The position of [SEP] token will not change during sampling, even though there exists token repetition in the generated sequences as shown in red.

\end{document}